\documentclass[runningheads]{llncs}
\usepackage{graphicx}

\usepackage{tikz}
\usepackage{comment}
\usepackage{amsmath,amssymb} 
\usepackage{color}

\usepackage{bbm}
\usepackage{graphicx}
\usepackage{booktabs}
\usepackage{enumitem}
\usepackage{algorithm}
\usepackage{algorithmic}
\usepackage{arydshln}
\usepackage{multirow}
\usepackage{tablefootnote}
\usepackage{threeparttable}
\usepackage{subcaption}
\usepackage{xspace}

\usepackage{hyperref}
\hypersetup{colorlinks=true,urlcolor=blue}
\definecolor{citecolor}{HTML}{0071bc}

\hypersetup{
  colorlinks,
  citecolor=citecolor,
  linkcolor=red
}

\newlength\savewidth\newcommand\shline{\noalign{\global\savewidth\arrayrulewidth
  \global\arrayrulewidth 1pt}\hline\noalign{\global\arrayrulewidth\savewidth}}

 \pdfoutput=1

  \makeatletter
\def\@fnsymbol#1{\ensuremath{\ifcase#1\or \dagger\or \ddagger\or
   \mathsection\or \mathparagraph\or \|\or **\or \dagger\dagger
   \or \ddagger\ddagger \else\@ctrerr\fi}}
\makeatother

\usepackage[accsupp]{axessibility}  

\makeatletter
\DeclareRobustCommand\onedot{\futurelet\@let@token\@onedot}
\def\@onedot{\ifx\@let@token.\else.\null\fi\xspace}

\def\eg{\emph{e.g}\onedot} 
\def\ie{\emph{i.e}\onedot} 
 
 \def\vs{\emph{vs}\onedot}

\makeatother

\def\0{{\bf 0}}
\def\1{{\bf 1}}

\def\hy{\hat{y}}

\def\hb{\hat{b}}

\def\citep{\cite}
\def\citet{\cite}

\begin{document}

\pagestyle{headings}
\mainmatter

\title{An Efficient Spatio-Temporal Pyramid Transformer for Action Detection} 


\titlerunning{An Efficient STPT for Action Detection}
%
\author{Yuetian Weng\inst{1} 
\and
Zizheng Pan\inst{1} 
\and
Mingfei Han\inst{1,2} 
\and
\\
Xiaojun Chang\inst{2,3} 
\and
Bohan Zhuang\inst{1}\thanks{Corresponding author.}} 
\authorrunning{Y. Weng, Z. Pan, M. Han, X. Chang and B. Zhuang}
%
\institute{Data Science \& AI, Monash University \and
ReLER Lab, AAII, University of Technology Sydney \and
School of Computing Technologies, RMIT University
\email{\{yuetian.weng,zizheng.pan,bohan.zhuang\}@monash.edu, hmf282@gmail.com, xiaojun.chang@uts.edu.au}}

\maketitle

\begin{abstract}

The task of action detection aims at deducing both the action category and localization of the start and end moment for each action instance in a long, untrimmed video. While vision Transformers have driven the recent advances in video understanding, it is non-trivial to design an efficient architecture for action detection due to the prohibitively expensive self-attentions over a long sequence of video clips. 
To this end, we present an efficient hierarchical Spatio-Temporal Pyramid Transformer (STPT) for action detection, building upon the fact that the early self-attention layers in Transformers still focus on local patterns. Specifically, we propose to use local window attention to encode rich local spatio-temporal representations in the early stages while applying global attention modules to capture long-term space-time dependencies in the later stages.
In this way, our STPT can encode both locality and dependency with largely reduced redundancy, delivering a promising trade-off between accuracy and efficiency.
For example, with only RGB input, the proposed STPT achieves 53.6\% mAP on THUMOS14, surpassing I3D+AFSD RGB model by over 10\% and performing favorably against state-of-the-art AFSD that uses additional flow features
with 31\% fewer GFLOPs, which serves as an effective and efficient end-to-end Transformer-based framework for action detection. 
\end{abstract}

\section{Introduction}
\label{sec:intro}

Action detection in lengthy, real-world videos is one of the crucial tasks in many video analysis applications, \eg, sports analysis, autonomous driving. 
Action detection aims to localize and classify the action instances appearing in untrimmed videos, which essentially depends on learning strong spatio-temporal representations from videos. 

To date, the majority of action detection methods~\cite{xu2017r_RC3D,li2020graph_AGCN,lin2018bsn,tan2021relaxed_RTDNet,chang2019mmvg} are driven by 3D convolutional neural networks (CNNs), \eg, C3D~\cite{7410867_C3D}, I3D~\cite{8099985_I3D}, to encode video segment features from video RGB frames and optical flows~\cite{zach2007duality}. 3D convolution is compact and effective to aggregate contextual pixels within a small 3D region, \eg, $3\times3\times3$, and thus reduce the spatio-temporal redundancy. However, the limited receptive field hinders the CNN-based models to capture long-term spatio-temporal dependencies. Alternatively, vision Transformers (ViTs) have shown the advantage~\cite{raghu2021vision} of capturing global dependencies via the self-attention mechanism in many computer vision tasks, such as image classification~\cite{dosovitskiy2020image,touvron2021training_deit,Pan_2021_ICCV} and video action recognition~\cite{bertasius2021space_TimeSformer,Zhang_2021_ICCV_vidtr,Fan_2021_ICCV_MViT,han2022dualai}. Hierarchical ViTs~\cite{Fan_2021_ICCV_MViT} divide Transformer blocks into several stages and progressively reduce the spatial size of feature maps when the network goes deeper. However, the high-resolution feature maps of video clips in the early stages result in overlong token sequences. For instance, given an input video clip with $256\times96\times96$ RGB frames, the feature maps after the initial embedding layer requires more than 1000G Floating-point Operations (FLOPs) for a standard multi-head self-attention layer, which is impractical to train or evaluate.
Therefore, how to efficiently handle spatio-temporal dependencies across overlong video frames is a fundamental challenge for action detection.

\begin{figure}[t]
\centering
\includegraphics[width=\textwidth]{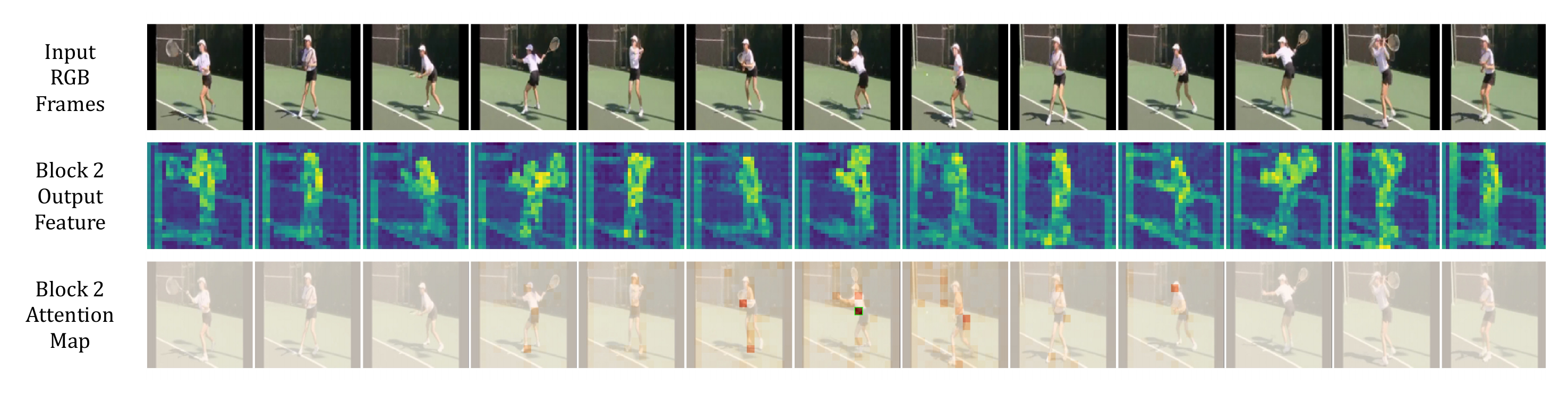}
\caption{Visualization of MViT~\cite{Fan_2021_ICCV_MViT}. We show the sampled input RGB frames, encoded feature maps and attention maps from the $2^{nd}$ block of MViT. We find that such global attention in the early stages actually encodes local visual patterns in each RGB frame, but it is relatively redundant in successive frames. For illustration, a target query token in the middle frame (green anchor) only attends to its nearby tokens in adjacent frames while rarely interacting with tokens in distant frames (filled with red, the more darker the color, the higher the attention score). Therefore, attending to all spatio-temporal tokens leads to huge redundancy in encoding such local patterns.}
\label{fig:attention_vis}
\end{figure}

In this paper, we present an efficient Spatio-Temporal Pyramid Transformer (STPT) to tackle both spatio-temporal redundancy and long-range dependency, as illustrated in Fig.~\ref{fig:architecture}. 
Specifically, we propose a Local Spatio-Temporal Attention blocks (LSTA) to capture local patterns in the early stages while introducing Global Spatio-Temporal Attention blocks (GSTA) to handle long-range spatio-temporal relationships in the later stages. 
The motivation of this design comes from two aspects. First, considering the spatial dimension in videos, previous studies in CNNs and ViTs~\cite{Hou_2017_CVPR,wu2019cascaded,raghu2021vision} have shown that shallow layers tend to capture local patterns in images (\eg, texture, edges) while deep layers tend to learn high-level semantics or capture long-range dependencies. Besides, target motions across adjacent frames are subtle, which implies large temporal redundancy when encoding video representations~\cite{feichtenhofer2019slowfast,wang2021adaptive,li2022uniformer}.
Therefore, it naturally gives rise to the question of whether applying global attention at the early stages to encode spatio-temporal representations is necessary.
Second, we empirically observe that heavy spatio-temporal redundancy exists in the shallow stages of current video Transformers~\cite{sharir2021image_16,bertasius2021space_TimeSformer,Arnab_2021_ICCV_ViViT,Zhang_2021_ICCV_vidtr,bulat2021space_SMAVT}. We take the $2^{nd}$ block of MViT \cite{Fan_2021_ICCV_MViT} as an example and visualize its output features as well as the attention maps in Fig.~\ref{fig:attention_vis}. We observe that the self-attention in the shallow layers mainly focuses on neighboring tokens in a small spatial area and adjacent frames, rarely attending to other tokens in distant frames. 
Hence, aggregating all the visual tokens via self-attention in the early stages can bring noises to the informative local representations and incurs huge computational redundancy. 
By leveraging LSTA in the early stages, STPT significantly alleviates spatio-temporal redundancy and inherently benefits spatio-temporal representation learning, as target motions are highly correlated in a local spatial region and temporally subtle across adjacent frames.

To encourage locality inductive bias, recent studies propose to combine convolutions~\cite{wu2021cvt}, MLPs~\cite{pan2021less_LIT} with Transformers or restrict self-attention within local windows~\cite{liu2021swin}, achieving favorable performance with reduced computational complexity. Moreover, from the theoretical perspective, locality inductive bias suppresses the negative Hessian eigenvalues, thus assisting in optimization by convexifying the loss landscape~\cite{park2022how}. 
Different from these methods, we are the pioneering work to build a pure Transformer model that encodes both compact spatial and temporal locality while preserving the long-range dependency for the action detection task. 

Finally, the proposed efficient Transformer is equipped with a temporal feature pyramid network (TFPN) to progressively reduce the spatial and temporal dimension and enlarge the receptive field into different scales. The multi-scale spatio-temporal feature representations are further utilized to predict the temporal boundaries and categories via an anchor-free prediction and refinement module. 

In summary, our contributions are in three folds: 
\begin{itemize}[leftmargin=*]
\item 
We propose an efficient and effective Spatio-Temporal Pyramid Transformer (STPT) for action detection, which reduces the huge computational cost and redundancy while capturing long-range dependency in spatio-temporal representation learning.

\item 
We devise local window attention to enhance local representations while reducing the spatio-temporal redundancy in shallow layers and retain the long-range dependency in deep layers with global self-attentions, achieving a favourable balance between efficiency and effectiveness.

\item
Finally, we conduct extensive experiments on standard benchmarks, \ie, THUMOS14 and ActivityNet 1.3, by using pure RGB frame input. Compared with the methods that combining additional flow features, our STPT achieves state-of-the-art results with reduced computational complexity, which makes a substantial stride for Transformer on the task of video action detection. 

\end{itemize}

\section{Related Work}

\subsection{Action Detection}

Action detection aims at localizing the temporal boundaries of human activities in untrimmed videos and classifying the action categories~\cite{vahdani2021deep_survey}. Most existing works~\cite{gao2017turn,li2020graph_AGCN,lin2018bsn} utilize
CNN-based models~\cite{7410867_C3D,8099985_I3D,qiu2017learning_P3D,8454294_TS} pretrained on large-scale datasets (\eg, Kinetics400~\cite{8099985_I3D}) to extract spatio-temporal representations from a stack of RGB frames and/or optical flow frames. Anchor-based
methods~\cite{gao2017turn,xu2017r_RC3D,chao2018rethinking_TAL} retrieve fine-grained proposals by adjusting pre-defined multi-scale anchors while actionness-guided methods~\cite{lin2018bsn,lin2019bmn,zhao2017temporal_SSN} instead learn the boundary confidence or actionness scores at all the temporal positions of the input video, which are matched and served as proposal candidates. 
Another line of research resorts to multi-scale towers~\cite{chao2018rethinking_TAL,9102850_TSA_Net} or temporal feature pyramids~\cite{lin2017single,zhang2018s3d,liu2020progressive_PBRNet} to tackle the variation of action duration, utilizing high-resolution feature maps for short actions and feature maps with large receptive field for long actions, respectively. Recently, a new anchor-free detector~\cite{lin2021learning_AFSD} directly predicts the distance to the action boundaries and the action category for each frame. However, the local receptive field of 3D convolutions leads to the loss of temporal dependencies on untrimmed videos. 
To capture the action dependencies across frames, prior works introduce graph models~\cite{zeng2019graph_PGCN,li2020graph_AGCN,bai2020boundary,Xu_2020_CVPR_GTAD,zhao2021video_VSGN}, RNNs~\cite{buch2017sst,buch2019end,yeung2016end}, and temporal Transformers~\cite{nawhal2021activity_AGT,tan2021relaxed_RTDNet,chang2021augmented_ATAG} to capture these temporal relationships. However, the aforementioned methods rely on pre-extracted features from 3D convolution backbones and use head-only learning manner.
In contrast, our STPT devises a pure Transformer model for efficiently and effectively learning spatio-temporal representations in an end-to-end manner, which encodes local patterns and global dependency via flexible token affinity learning in shallow and deep layers, respectively.

\subsection{Video Transformers}

ViTs are pushing the boundaries of recent video understanding research.
In particular, VTN~\cite{neimark2021video_VTN}, LightVideoFormer~\cite{koot2021videolightformer} and STAM~\cite{sharir2021image_16} introduce temporal Transformers to encode inter-frame relationships over the extracted image-level feature maps. ViViT~\cite{Arnab_2021_ICCV_ViViT},  TimeSformer~\cite{bertasius2021space_TimeSformer} and VidTr~\cite{Zhang_2021_ICCV_vidtr} propose to factorize along spatial and temporal dimensions on the granularity of encoder, attention block or dot-product computation. Similarly, SCT~\cite{zha2021shifted_SCT} proposes image chunk attention and shifted attention to model spatial and temporal relationships respectively. SMAVT~\cite{bulat2021space_SMAVT} aggregates information from tokens located at the same spatial location within a local temporal window, while SIFAR~\cite{fan2021image_SIFAR} turns spatio-temporal patterns in video into purely spatial patterns in images, showing an image classifier can undertake the task of video understanding. However, these studies lack hierarchical structure or model spatio-temporal dependencies separately, which may not be sufficient for the task of action detection. Targeting on these issues, MViT~\cite{Fan_2021_ICCV_MViT} presents a hierarchical Transformer to progressively shrink the spatio-temporal resolution of feature maps while expanding the channel as the network goes deeper. VideoSwin~\cite{liu2021video_videoSwin} proposes shifted window attention to limit the computation within a small local window, while Uniformer~\cite{li2022uniformer} unifies the spatio-temporal MobileNet block and self-attention and proposes an alternative multi-head relation aggregator.
Different from the above methods, our model is purely Transformer based and jointly learns spatio-temporal representation. By flexibly involving locality constraint in early stages and data-specific global self-attentions in later stages, our model well addresses the challenges of spatio-temporal redundancy and dependency for action detection. 

\begin{figure}[t]
\centering
\includegraphics[width=\textwidth]{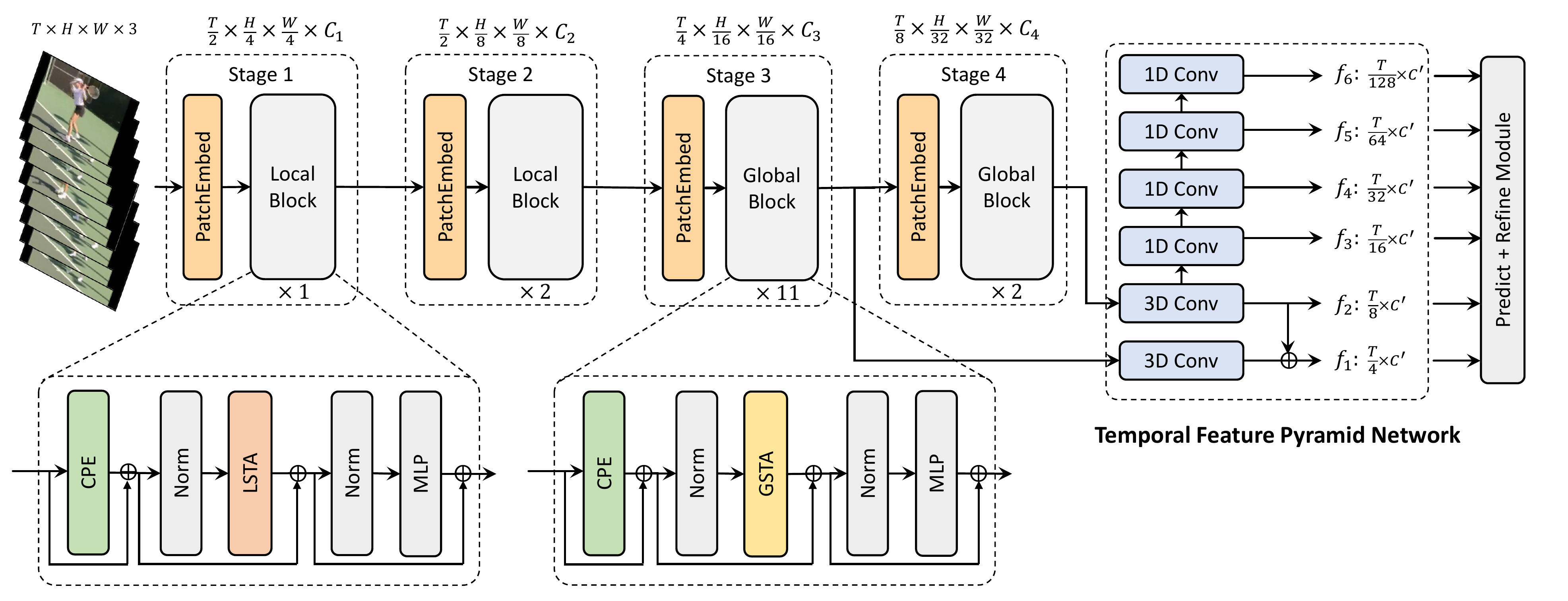}
\caption{Overall architecture of the proposed spatio-temporal pyramid Transformer. Given an input video $X_{in}$, our STPT is utilized to encode spatio-temporal representations and detect existing action instances. In the shallow layers, local blocks that constrain self-attention into local 3D windows tend to encode local patterns while reducing spatio-temporal redundancy. In deeper layers, global blocks retain long-range dependency over the token sequence. Features from the last stages are fed to the temporal feature pyramid network to predict the temporal boundaries and action categories via an anchor-free prediction and refinement module. Please refer to Sec.~\ref{sec:arc} for additional details.}
\label{fig:architecture}
\end{figure}

\section{Method}
\subsection{Overall Architecture}
\label{sec:arc}
The overall architecture of STPT is illustrated in Fig.~\ref{fig:architecture}. Let $X_{in} \in \mathbb {R}^{T\times H\times W\times3}$ be an input video clip, where $T$, $H$ and $W$ represent the number, height and width of RGB frames, respectively. First, we divide the frame volume $X_{in}$ into a series of overlapping 3D cubes, with the size of $3\times7 \times7$.
Then we exploit a 3D PatchEmbed module to aggregate contextual pixels from each cube and project each cube into dimension of 96, serving as the initial input for the subsequent pipeline.
In practice, the 3D PatchEmbed module is a $3\times7\times7$ depth-wise convolution with stride $2\times4\times4$ and zero paddings. As a result, the input tensor $X_{in}$ is downsampled to $\frac{T}{2}\times \frac{H}{4} \times \frac{W}{4}$. 

Next, we divide the entire backbone into 4 stages. Letting $s\in[1,2,3,4]$ be the index of each stage, we employ $L_s\in[1,2,11,2]$ blocks at each stage, in which STPT Local Blocks are used to encode local spatio-temporal representations in the first two stages, and STPT Global Block are used to tackle long-term dependencies in the later stages. At the same time, from the $2^{nd}$ stage, the spatial and temporal dimensions are gradually downsampled by the 3D PatchEmbed module. Following \cite{Fan_2021_ICCV_MViT}, by increasing the output dimension of the final MLP layer in the previous stage, the channel dimensions are gradually expanded before the transition to the next stage. The feature maps with different resolutions from the last two stages are then fed to the temporal feature pyramid network (TFPN) to obtain multiple resolution feature maps. Finally, the prediction and refinement modules are used to predict the start frame, end frame and the category of each anchor point on the multi-scale feature maps. The detailed architecture specifications are provided in Table \ref{tab:archi_specification}.

\begin{table}[t]
\begin{center}
\caption{Architecture specification of STPT. For the $s$-th stage, we denote $P_s$ as the patch size, $S_s$ as the kernel stride and $R_s$ as the reduction ratio, where each dimension corresponding to the temporal size, height and width, respectively. $L_s$ and $C_s$ refer to the the number of blocks and the channel dimension at the $s$-th stage, respectively. Furthermore, we denote $W_{sl}$ as the window size at the $l$-th block in the $s$-th stage. We adopt an expanding ratio of 4 for all MLP layers in each block.}
\label{tab:archi_specification}
\begin{tabular}{c|c|c|c}
                        & Output Size                            & Layer Name        & STPT                                                                                                                           \\ \shline
\multirow{4}{*}{Stage1} & \multirow{4}{*}{$\frac{T}{2}\times\frac{H}{4}\times\frac{W}{4}$} & PatchEmbed        & \begin{tabular}[c]{@{}c@{}}$P_1=3\times7\times7$\\ $S_1=2\times4\times4$\\ $C_1=96$\end{tabular}                               \\ \cline{3-4} 
                        &                                        & Local Block  & \begin{tabular}[c]{@{}c@{}}$L_1=1$\\ $W_{11}=8\times8\times8$\\ $R_1=2\times8\times8$\end{tabular}                                \\ \hline
\multirow{4}{*}{Stage2} & \multirow{4}{*}{$\frac{T}{2}\times\frac{H}{8}\times\frac{W}{8}$} & PatchEmbed        & \begin{tabular}[c]{@{}c@{}}$P_2=3\times3\times3$\\ $S_2=1\times2\times2$\\ $C_2=192$\end{tabular}                              \\ \cline{3-4} 
                        &                                        & Local Block  & \begin{tabular}[c]{@{}c@{}}$L_2=2$\\ $W_{21}=8\times6\times6$\\ $W_{22}=16\times4\times4$\\ $R_2=2\times2\times2$\end{tabular} \\ \hline
\multirow{4}{*}{Stage3} & \multirow{4}{*}{$\frac{T}{4}\times\frac{H}{16}\times\frac{W}{16}$}    & PatchEmbed        & \begin{tabular}[c]{@{}c@{}}$P_3=3\times3\times3$\\ $S_3=2\times2\times2$\\ $C_3=384$\end{tabular}                              \\ \cline{3-4} 
                        &                                        & Global Block & \begin{tabular}[c]{@{}c@{}}$L_3=11$\\ $R_3=2\times2\times2$\end{tabular}                                                       \\ \hline
\multirow{4}{*}{Stage4} & \multirow{4}{*}{$\frac{T}{8}\times\frac{H}{32}\times\frac{W}{32}$}    & PatchEmbed        & \begin{tabular}[c]{@{}c@{}}$P_4=3\times3\times3$\\ $S_4=2\times2\times2$\\ $C_4=768$\end{tabular}                              \\ \cline{3-4} 
                        &                                        & Global Block & \begin{tabular}[c]{@{}c@{}}$L_4=2$\\ $R_4=1\times1\times1$\end{tabular}                                                        \\ 
\end{tabular}
\end{center}
\end{table}

\subsection{Block Design in STPT}
\label{sec:block}

Formally, each STPT Block consists of three key modules: Conditional Positional Encoding (CPE), Multi-Head Spatio-Temporal Attention (MSTA), and multi-layer perceptron (MLP).
Formally, letting $X_{l-1}$ be the input of the $l$-th block, each block can be formulated as
\begin{align}
  X_{l-1}  & = {\rm{CPE}}(X_{l-1}) + X_{l-1} \; \\
  X'_{l-1} & = {\rm{MSTA}}({\rm{LN}}(X_{l-1})) + X_{l-1} \; \\
  X_{l}  & = {\rm{MLP}}({\rm{LN}}(X'_{l-1})) + X'_{l-1}
\end{align}
where $\rm{LN}(\cdot)$ indicates the layer normalization \cite{ba2016layer} and MLP consists of two FC layers with GELU~\cite{gelu} non-linearity in between. Specifically,
we first leverage CPE to integrate the spatio-temporal position information into each token. 
Then the MSTA module, which can be the Local spatio-temporal Attention (LSTA) or Global spatio-temporal Attention (GSTA), aggregates each token with its contextual tokens, followed by an MLP to perform channel mixing. 
As discussed above, we encode fine and local spatial-temporal representations in the early stages using LSTA and high-level semantics with long-term dependencies in the deep layers using GSTA, respectively. In the next, we elaborate the design for each module. 

\noindent\textbf{Conditional positional encoding.}
Since actions in videos are both spatial and temporal variant, we need to explicitly encode position information for all the visual tokens. 
Previous works~\cite{Fan_2021_ICCV_MViT,Arnab_2021_ICCV_ViViT,bertasius2021space_TimeSformer} commonly adopt absolute positional encodings~\cite{dosovitskiy2020image}. However, the length of token sequences is much longer for action detection compared to the one pretrained for action recognition tasks (\eg, $128\times24\times24$ vs. $8\times56\times56$), which makes it difficult to utilize a fixed absolute positional encoding from a pretrained model.
To tackle these problems, we make the spatio-temporal positional embedding conditioned on input features and extend the CPE proposed in \cite{chu2021conditional} to the video domain, which can be formulated as 
\begin{align}
{\rm{CPE}}(X) = {\rm{DWConv}}(X),
\end{align}
where $\rm{DWConv}$ refers to a 3D depth-wise convolution with zero paddings. Previous works~\cite{chu2021conditional,Islam*2020How} have shown that tokens on the borders can be aware of their absolute positions when using convolutional layers with zero paddings. Therefore, the absolute position for each token can be encoded by progressively sliding convolutional kernels on the feature maps, justifying the design of our proposed CPE for introducing positional information into input features.

\noindent\textbf{Local spatio-temporal attention.}
As discussed in Sec. \ref{sec:intro}, global attention is redundant to encode local patterns in the shallow layers, thus leading to high computational cost owing to the high-resolution feature maps. Given an embedding $X \in \mathbb {R}^{T\times H\times W\times d}$ with channel dimension $d$, the complexity of self-attention 
is $\mathcal{O}(T^2H^2W^2d)$ \cite{vaswani2017attention}. Here, we propose to replace self-attention with LSTA to alleviate the redundancy. 

Following the design of multi-head self-attention, LSTA first projects $X$ into query $Q\in\mathbb{R}^{T\times H\times W \times d}$, key $K\in\mathbb{R}^{T\times H\times W \times d}$, and value $V\in\mathbb{R}^{T\times H\times W \times d}$ with linear transformations. 
For each tensor, we evenly divide it into $w_1 \times w_2 \times w_3$ partitions (sub-windows). Without loss of generality, we assume $T \% w_1 = 0$, $H \% w_2 = 0$ and $W \% w_3 = 0$, and thus each sub-window contains $\frac{THW}{w_1w_2w_3}$ tokens. 
We force each query token only attends to tokens within the same local 3D window, which helps encode local patterns via joint spatial-temporal relation aggregation and reduce the computational redundancy in the early stages.

To further improve the efficiency, we follow~\cite{wang2021pyramid_PvT,Fan_2021_ICCV_MViT} to reduce both the spatial and temporal resolution of the keys and values within each local window, \ie, $\bar{K} = R_K(K) \in \mathbb{R}^{T'\times H' \times W' \times d}$ and $\bar{V} = R_V(V) \in \mathbb{R}^{T'\times H' \times W' \times d}$, where $\bar{K}$, $\bar{V}$ are resolution-reduced keys and values, and $R_K$, $R_V$ denote two independent reduction operations (\eg, depth-wise convolution). $T'$, $H'$ and $W'$ are the reduced temporal dimension, height and width. Thus, each sub-window of $\bar{K}$ and $\bar{V}$ contains $\frac{T'H'W'}{w_1 w_2 w_3}$ tokens after being divided into ${w_1 \times w_2 \times w_3}$ partitions. 
Specifically, the computational cost for each sub-window becomes $\mathcal{O}(\frac{THW}{w_1 w_2 w_3} \times \frac{T'H'W'}{w_1 w_2 w_3}\times d)$, and the total cost of LSTA is $\mathcal{O}(\frac{T'H'W'}{w_1 w_2 w_3}\times THWd)$, which is significantly efficient when $\frac{T'}{w_1} \ll T$, $\frac{H'}{w_2} \ll H$ and $\frac{W'}{w_3} \ll W$ and grows linearly with $THW$ if $\frac{T'}{w_1}$, $\frac{H'}{w_2}$ and $\frac{W'}{w_3}$ are fixed. 

By applying LSTA in the early stages, the model significantly alleviates spatio-temporal redundancy and efficiently encodes local spatio-temporal representations. 

\noindent\textbf{Global spatio-temporal attention.}
To capture long-term dependencies, we employ GSTA in the deep layers. 
For more efficient aggregation, GSTA also uses the feature maps with reduced spatio-temporal resolution as the keys and values in the self-attention operations. 
Given a query token, GSTA compares it with all the tokens for aggregation. 
In this way, we ensure that the model captures global dependencies in the last stages. By combining LSTA in the shallow layers, the model forms an efficient and effective way of learning the spatio-temporal representations for action detection.

\noindent\textbf{Relation to existing video Transformers.}
While \cite{bertasius2021space_TimeSformer,Arnab_2021_ICCV_ViViT} are based on space-time attention factorization, our method can encode the target motions by jointly aggregating spatio-temporal relations, without loss of spatio-temporal correspondence. 
Compared with MViT~\cite{Fan_2021_ICCV_MViT} which entirely utilizes global self-attentions, our model can resolve the long-range dependency while simultaneously reducing the local spatio-temporal redundancy. By removing the local redundancy in the early stages, our model outperforms MViT with a lower computational cost. Moreover, different from the spatio-temporal MobileNet block used in Uniformer~\cite{li2022uniformer}, our LSTA is data-dependent~\cite{park2022how,han2022on} and flexible in terms of window size without introducing extra parameters, while the kernel parameters are fixed for the 3D convolutions in the spatio-temporal MobileNet block. 
Different from VideoSwin~\cite{liu2021video_videoSwin}, we do not use shifted window mechanism to get a trade-off between locality and dependency. 
Compared with~\cite{chen2022regionvit,chu2021twins,liang2021dualformer} which alternatively process local and global information within each block, we encode local and global spatio-temporal representations in the shallow and deep layers separately, tackling both redundancy and dependency in a concise manner.

\subsection{Temporal Feature Pyramid}
Given an untrimmed video, action detection aims to find the temporal boundaries and categories of action instances, with annotation denoted by $\{\psi_n=(t_{n}^{s}, t_{n}^{e}, c_n)\}_{n=1}^{N}$, where $N$ is the number of action instances. For the $n$-th action instance $\psi_n$, $t_{n}^{s}$, $t_{n}^{e}$ and $c_n$ refer to the start time, end time and action label, respectively.  
As shown in Fig.~\ref{fig:architecture}, we first forward the video input $X_{in}$ into the backbone to encode the spatio-temporal representations. The 3D feature maps extracted from the last two stages are then fed to TFPN to obtain multi-scale temporal feature maps. The motivation comes from the fact that multi-scale feature maps contribute to tackle the variation of action duration~\cite{chao2018rethinking_TAL,9102850_TSA_Net,wu2021towards}. 
Specifically, we construct an $M$-level temporal feature pyramid $\{f_m\}_{m=1}^{M}$, where $f_m\in \mathbb{R}^{T_m\times C'}$ and $T_m$ is the temporal dimension of the $m$-th level.
The TFPN contains two 3D convolution layers followed by four 1D convolutional layers to progressively forms a featural hierarchy.

After obtaining the temporal feature pyramid, an anchor-free prediction and refinement module as in \cite{lin2021learning_AFSD} is utilized to predict the boundary distances and class scores at each location $i$ on $f_m$. Concretely, a two-branch tower including several temporal convolutional layers is employed to map $f_m$ into two latent representations. 
The latent representations are then processed by a classification head and a localization head to get the class label $\hy_i^C$ and boundary distances $(\hb_i^s, \hb_i^e)$ for each location $i$, respectively. To improve the confidence of the predictions, we further adjust the boundary distances with features extracted from a small region at the coarse boundary predicted above to obtain the modified offset as $(\Delta \hb_i^s, \Delta \hb_i^e)$, and the refinement action category label as $\hy_i^R$. 
To obtain high quality proposals, we additionally predict the quality confidence $\eta$ following~\cite{zhao2017temporal_SSN}. Formally, for the $i$-th temporal location in the $m$-th TFPN layer, the final predicted start time $\hat{t}_{m,i}^{s}$, end time $\hat{t}_{m,i}^{e}$ and class label $\hy_{m,i}$ can be formulated in the following form:
\begin{align}
    \hat{t}_{m,i}^{s} & = \hb_{m,i}^{s} + \frac{1}{2}(\hb_{m,i}^{e}-\hb_{m,i}^{s})\Delta \hb_i^s, \; \\
    \hat{t}_{m,i}^{e} & = \hb_{m,i}^{e} + \frac{1}{2}(\hb_{m,i}^{e}-\hb_{m,i}^{s})\Delta \hb_i^e, \; \\
    \hy_{m,i} & = \frac{1}{2}(\hy_{m,i}^C+\hy_{m,i}^R)\eta_{m,i}.
\end{align}

In the training process, we use a multi-task loss function based on the output of coarse and refined predictions, which can be formulated as 
\begin{align}
    \mathcal{L} = \lambda_{cls} \mathcal{L}_{cls} + \lambda_{loc} \mathcal{L}_{loc} + \lambda_{q} \mathcal{L}_q,
\end{align}
where $\mathcal{L}_{cls}$, $\mathcal{L}_{loc}$ and $\mathcal{L}_{q}$ are losses corresponding to the classification, boundary regression and quality confidence prediction tasks, respectively, and $\lambda_{cls}$, $\lambda_{loc}$, $\lambda_{q}$ are hyperparameters to balance the contribution of each task to the total loss. For the classification task, we use focal loss~\cite{lin2017focal} between the predicted action scores from both prediction and refinement modules and the ground-truth categories, \ie, $\mathcal{L}_{cls}=\mathcal{L}^C_{focal}+\mathcal{L}^R_{focal}$. The localization loss includes tIoU (temporal Interaction over Union) loss for the predicted coarse boundaries and L1 loss for refined offsets, respectively, \ie, $\mathcal{L}_{loc}=\mathcal{L}_{tIoU}^C+\mathcal{L}_{L1}^R$. For the quality prediction task, $\mathcal{L}_q$ is computed in the same way as in \cite{lin2021learning_AFSD}. 

\section{Experiments}
\subsection{Datasets and Settings}

\noindent\textbf{Datasets.}
We present our experimental results on the commonly-used benchmarks THUMOS14~\cite{THUMOS14} and ActivityNet 1.3~\cite{caba2015activitynet}. THUMOS14 dataset is composed of 413 temporally annotated untrimmed videos with 20 action categories. We use the 200 videos in the validation set for training and evaluate our method on the 213 videos in the test set. ActivityNet 1.3 is a large-scale action understanding dataset for action recognition, action detection, proposal generation and dense captioning tasks, which contains 19,994 temporally labeled untrimmed videos with 200 action categories. We follow the former setting~\cite{lin2018bsn} to split this dataset into training, validation and testing sets based on the proportion of 2:1:1.

\noindent\textbf{Metrics.} We adopt mean Average Precision (mAP) at certain tIoU thresholds as the evaluation metric. On THUMOS14, we use tIoU thresholds \{0.3, 0.4, 0.5, 0.6, 0.7\}; on ActivityNet 1.3, we choose 10 values in the range of [0.5, 0.95] with a step size of 0.05 as tIoU thresholds following the official evaluation API. 

\noindent\textbf{Implementation details.}
We build our STPT based on MViT~\cite{Fan_2021_ICCV_MViT}. The pipeline and  architecture specifications have shown in Fig.~\ref{fig:architecture} and Table~\ref{tab:archi_specification}, respectively. Follow common practice~\cite{lin2021learning_AFSD}, we train and evaluate our model in an end-to-end manner, which takes as input pure RGB frames without using additional optical flow features. On THUMOS14, we sample RGB frames at 10 frames per second (fps) and split a video into clips, where each clip contains 256 frames. Adjacent clips have a temporal overlap of 30 and 128 frames at training and testing, respectively. For ActivityNet 1.3, we sample a clip of 768 frames at dynamic fps for each video. We set the spatial resolution as $96\times96$ and use data augmentation including random crop and horizontal flipping in training. 
For a fair comparison, we pretrain all the models on Kinetics400~\cite{8099985_I3D} for 30 epochs under the same settings following~\cite{Fan_2021_ICCV_MViT}. 
Our model is trained for 16 epochs and 12 epochs on THUMOS14 and ActivityNet 1.3, respectively, using Adam~\cite{kingma2014adam} with a learning rate of $5\times10^{-6}$ for backbone and $1\times10^{-4}$ for other modules, and the weight decay is set to $1\times10^{-3}$ and $1\times10^{-4}$ for the two datasets. In post-processing, we apply soft-NMS~\cite{bodla2017soft_nms} to suppress redundant predictions, where the tIoU threshold is set to 0.5 for THUMOS14 and 0.85 for ActivityNet 1.3. $\lambda_{loc}$ is set to 10 for THUMOS14 and 1 for ActivityNet 1.3, and $\lambda_{cls}$, $\lambda_q$ is set to 1. 

\begin{table}[t]
\begin{center}
\caption{Performance comparison with state-of-the-art methods on THUMOS14, We measure the performance by mAP at different tIoU thresholds and average mAP in $[0.3 : 0.1 : 0.7]$. ``*'' indicates that the models are trained in an end-to-end manner. We measure the computational cost by GFLOPs based on a clip of $256\times96\times96$ frames for them. ``Flow'' indicates using optical flow features.}
\label{table:thumos14}
\begin{tabular*}{\textwidth}{c|c|c|c|@{\extracolsep{\fill}}cccccc}
Methods     & GFLOPs  & Backbone & Flow & 0.3  & 0.4  & 0.5  & 0.6  & 0.7  & Avg.   \\ 
\shline
BSN~\cite{lin2018bsn}        & 455.4      & TS             & \checkmark     & 53.5 & 45.0 & 36.9 & 28.4 & 20.0 & 36.8  \\
BMN~\cite{lin2019bmn}        & 455.4       & TS             & \checkmark     & 56.0 & 47.4 & 38.8 & 29.7 & 20.5 & 38.5  \\
G-TAD~\cite{Xu_2020_CVPR_GTAD}      & 444.2      & TSN             & \checkmark     & 54.5 & 47.6 & 40.2 & 30.8 & 23.4 & 39.3  \\
TAL~\cite{chao2018rethinking_TAL}        & 157.0      & I3D            & \checkmark     & 53.2 & 48.5 & 42.8 & 33.8    & 20.8    & 39.8     \\
A2Net~\cite{yang2020revisiting_A2Net}      & 157.0      & I3D            & \checkmark     & 58.6 & 54.1 & 45.5 & 32.5 & 17.2 & 41.6  \\
G-TAD+PGCN~\cite{zeng2019graph_PGCN} & 157.0      & I3D            & \checkmark     & 66.4 & 60.4 & 51.6 & 37.6 & 22.9 & 47.8  \\
BMN-CSA~\cite{sridhar2021class_CSA}    & 455.4      & TS             & \checkmark     & 64.4 & 58.0 & 49.2 & 38.2 & 27.8 & 47.5  \\
AFSD*~\cite{lin2021learning_AFSD}       & 162.2 & I3D            & \checkmark     & 67.3 & 62.4 & 55.5 & 43.7 & 31.1 & 52.0  \\
DCAN~\cite{2022dcan} & 444.2      & TSN            &  \checkmark    & 68.2 & 62.7 & 54.1 & 43.9 & \textbf{32.6} & 52.3  \\

\hline
R-C3D*~\cite{xu2017r_RC3D}      & 453.3      & C3D            &      & 44.8 & 35.6 & 28.9 & -    & -    & -     \\
GTAN~\cite{long2019gaussian}       & 107.0      & P3D            &      & 57.8 & 47.2 & 38.8 & -    & -    & -     \\
AFSD*~\cite{lin2021learning_AFSD}       & 84.4  & I3D            &      & -    & -    & 45.9 & 35.0 & 23.4 & 43.5  \\
BCNet+PGCN~\cite{yang2022temporal} & 81.8 & I3D &  & 69.8 & 62.9 & 52.0 & 39.8 & 24.0 & 49.7 \\
DaoTAD~\cite{wang2021rgb}     & 81.8      & I3D            &      & 62.8 & 59.5 & 53.8 & 43.6 & 30.1 & 50.0  \\ 

\hline
           & 167.6 & MViT~\cite{Fan_2021_ICCV_MViT}           &      & 68.0 & 62.5 & 54.2 & 43.6 & 30.6 & 51.8  \\
           & 120.9 & VideoSwin~\cite{liu2021video_videoSwin} &       & 
           69.5 & 64.1 & 54.7 & 42.6 & 27.7 & 51.7   \\
           & 116.1 & TimeSformer~\cite{bertasius2021space_TimeSformer}    &      & 67.6 & 61.9 & 53.0 & 41.9 & 27.9 & 50.5  \\
Ours*      & 111.2 & STPT           &      & \textbf{70.6} & \textbf{65.7} & \textbf{56.4} & \textbf{44.6} & 30.5 & \textbf{53.6}  \\
\end{tabular*}
\end{center}
\end{table}

\subsection{Main Results}

We compare our STPT with state-of-the-art approaches on the two datasets in Table~\ref{table:thumos14} and Table~\ref{table:anet}. We also report the backbone used by each method, \eg, I3D~\cite{8099985_I3D}, TS~\cite{simonyan2014two}, TSN~\cite{8454294_TS}, P3D~\cite{qiu2017learning_P3D}, and whether the optical flows are used. For models that are end-to-end trainable, we directly report the computational cost for the whole model. For methods utilizing pre-extracted features or pre-generated action proposals, we also calculate the computational cost for the offline feature extraction stage under identical input settings for fair comparison. 

On THUMOS14 dataset, our STPT, which only uses RGB frames, outperforms previous RGB models by a large margin and achieves comparable performance with methods using additional optical flows, reaching mAP 70.6\%, 65.7\%, 56.4\% at tIoU thresholds 0.3, 0.4, 0.5, respectively. 
Specifically, our STPT provides 53.6\%, a +10.1\% average mAP boost over AFSD RGB model under identical settings. Besides, our model also outperforms the two-stream AFSD at most tIoU thresholds with less computational cost (111.2 vs. 162.2 GFLOPs).

On ActivityNet, with significant reduction of computational cost, our STPT also achieves comparable performance with the existing RGB models. Specifically, STPT still obtains slightly better mAPs than AFSD RGB model on all thresholds with less computational cost (134.1 vs. 248.7 GFLOPs). Notably, most previous models use optical flows to enhance motion modeling. 
However, the adoption of an ensemble of flow features requires pre-extracting flow features using~\cite{zach2007duality}, which prevents these methods from end-to-end learning and also introduces huge computational cost. In contrast, our STPT can effectively encode the spatio-temporal representations from pure RGB frames, which is completely end-to-end trainable and computationally efficient.

It is worth to note that, compared with other representative video Transformers, our STPT achieves the best mAP on THUMOS14 and ActivityNet, which demonstrates the advantages of our architecture design that applying LSTA in shallow layers and GSTA in deeper layers in realizing the trade-off between locality and dependency. We also provide more comparison with other representative video Transformers in the supplementary.

\begin{table}[t]
\begin{center}
\caption{Action localization results on ActivityNet 1.3 (validation set), measured by mAP(\%) at different tIoU thresholds, and the average mAP in $[0.5 : 0.05 : 0.95]$. }
\label{table:anet}
\begin{tabular}{c|c|c|c|cccc} 
Models  & GFLOPs  & Backbone & Flow & 0.5  & 0.75 & 0.95 & Avg.  \\ 
\shline
TAL~\cite{chao2018rethinking_TAL}     & 471.3      & I3D            & \checkmark     & 38.2 & 18.3 & 1.3  & 20.2  \\
A2Net~\cite{yang2020revisiting_A2Net}   & 471.3      & I3D            & \checkmark     & 43.6 & 28.7 & 3.7  & 27.8  \\
BSN~\cite{lin2018bsn}     & 1367.0      & TS             & \checkmark     & 46.5 & 30.0 & 8.0  & 30.0  \\
BMN~\cite{lin2019bmn}     & 1367.0      & TS             & \checkmark     & 50.1 & 34.8 & 8.3  & 33.9  \\
G-TAD~\cite{Xu_2020_CVPR_GTAD}   & 1367.0      & TS            & \checkmark     & 50.4 & 34.6 & 9.0  & 34.1  \\
BMN-CSA~\cite{sridhar2021class_CSA} & 1367.0      & TS             & \checkmark     & 52.4 & 36.7 & 5.2  & 35.4  \\ 
AFSD*~\cite{lin2021learning_AFSD}    & 478.0  & I3D            & \checkmark     & 52.4 & 35.3 & 6.5  & 34.4  \\
DCAN~\cite{2022dcan} & 1367.0 & TS & \checkmark &  51.8 & 36.0 & 9.5 & 35.4 \\
\hline
R-C3D*~\cite{xu2017r_RC3D}   & 1360.0      & C3D            &      & 26.8 & -    & -    & 12.7  \\
GTAN~\cite{long2019gaussian}    & 320.0      & P3D            &      & 52.6 & 34.1 & 8.9  & 34.3  \\
AFSD*~\cite{lin2021learning_AFSD}    & 248.7 & I3D            &      & 50.5 & 33.4 & 6.5  & 32.9  \\
\hline
        & 172.4 & MVIT~\cite{Fan_2021_ICCV_MViT}           &      & 50.1 & 32.7 & 5.9  & 32.2  \\
        & 153.7 & VideoSwin~\cite{liu2021video_videoSwin} &       & 49.6                & 32.1 & 5.6 & 31.9 \\
        & 140.6 & TimeSformer~\cite{bertasius2021space_TimeSformer}    &      & 51.1 & 33.3 & 6.0  & 33.1  \\
Ours*    & 134.1 & STPT           &      & 51.4 & 33.7 & 6.8  & 33.4  \\
\end{tabular}
\end{center}
\end{table}

\subsection{Ablation Study}

\noindent\textbf{Effect of LSTA.} As discussed in Sec.~\ref{sec:intro}, MViT~\cite{Fan_2021_ICCV_MViT} suffers from heavy spatio-temporal redundancy in the shallow layers. We also investigate the computational cost of each layer and find that the heavy computation is caused by the global attention in the first two stages. To this end, we replace global attention with our efficient LSTA in the first two stages. 
As shown in Table~\ref{table:thumos14}, our STPT improves baseline MViT, which actually uses global self-attention in all the stages, by $1.4\%$ on average mAP while reducing FLOPs by $55.4$G on THUMOS14. 
Furthermore, we also compare our LSTA with factorized space-time attention in~\cite{bertasius2021space_TimeSformer,Arnab_2021_ICCV_ViViT}. As shown in Table~\ref{table:thumos14} and Table~\ref{table:anet}, our LSTA leads to higher mAP scores than other attention designs with less computational cost, indicating the effectiveness and efficiency of the proposed module. 

\noindent\textbf{Effect of the architecture design.} To explore the effect of our architecture design principle in STPT, we investigate all the possible combinations of LSTA (L) and GSTA (G). As shown in Table~\ref{table:archi_des}, when only using LSTA, the computational cost is light (LLLL). However, the mAP drops dramatically, since the network lacks the capacity of learning global dependency without GSTA. 
A significant improvement can be observed when replacing LSTA with GSTA in the $3^{rd}$ stage, which indicates the importance of learning global dependency in the deeper layers for action detection. 
However, when applying GSTA in all stages (GGGG), the model leads to worse results and introduces heavy computational overhead (111.2G vs 166.6G). The main reason is that, without locality constraints, the model cannot extract detailed spatio-temporal patterns in the early stages.
In our experiments, we choose LSTA and GSTA in the first two stages and the last two stages respectively to achieve a preferable balance between efficiency and effectiveness. 

\begin{table}[t!]
\begin{center}
\caption{Effect of our architecture design principle. We evaluate the performance of several combinations of blocks in terms of mAP. L/G refers to LSTA/GSTA blocks used in each stage. All models are equipped with CPE. The number of temporal tokens is set to 128, and the window size of temporal dimension for LSTA is set to 8. }
\label{table:archi_des}
\begin{tabular*}{0.8\textwidth}{c|c|@{\extracolsep{\fill}}cccccc}
Type & GFLOPs & 0.3  & 0.4  & 0.5  & 0.6  & 0.7  & Avg.  \\ \shline
LLLL & 101.8  & 64.7 & 59.4 & 50.8 & 40.0 & 26.5 & 48.3 \\ \hline
LLLG & 102.7  & 67.6 & 62.1 & 54.2 & 42.2 & 29.4 & 51.3 \\ \hline
LGGG & 151.4  & 69.4 & 63.5 & 55.0 & 42.9 & 29.2 & 52.0 \\ \hline
GGGG & 167.6  & 68.0 & 62.5 & 54.2 & 43.6 & 30.6 & 51.8 \\ \hline
LLGG & 111.4  & 69.1 & 63.7 & 55.2 & 44.2 & 29.3 & 52.3 \\
\end{tabular*}
\end{center}
\end{table}

\begin{table}[t!]
\begin{center}
\caption{Effect of window size in terms of temporal dimension on THUMOS. We compare the performance (in mAP) and computational cost (in GFLOPs) for different scales of local window in each LSTA block. The number of temporal tokens is set to 128. Experiments are conducted on models without using CPE.}
\label{table:window_size}
\begin{tabular*}{0.8\textwidth}{c|c|@{\extracolsep{\fill}}cccccc}
Window size    & GFLOPs & 0.3  & 0.4  & 0.5  & 0.6  & 0.7  & Avg. \\ \shline
{[}1,1,1{]}    & 110.4  & 67.0 & 61.5 & 53.6 & 41.3 & 28.3 & 50.3 \\ \hline
{[}4,4,4{]}    & 110.6  & 67.8 & 62.4 & 53.4 & 41.6 & 28.1 & 50.6 \\ \hline
{[}8,8,8{]}    & 110.8  & 67.6 & 62.5 & 53.1 & 42.1 & 29.1 & 50.9 \\ \hline
{[}8,8,16{]}   & 110.8  & 69.5 & 63.6 & 55.6 & 44.9 & 29.5 & 52.7 \\ \hline
{[}16,16,16{]} & 111.1  & 66.3 & 61.1 & 52.7 & 41.5 & 28.7 & 50.1 \\ 
\end{tabular*}
\end{center}
\end{table}

\noindent\textbf{Effect of the window size.} We compare the results of different window sizes in LSTA in terms of the temporal dimension. As shown in Table~\ref{table:window_size}, LSTA is beneficial from the suitable window size. 
Moreover, it becomes equivalent to encode spatial patterns without temporal information when using only [1,1,1] in LSTA, where the model can not capture the motion variation of local patterns in the shallow layers, leading to performance drop for action detection. However, the setting of [16,16,16] leads to 0.5\% average mAP drop compared to [8,8,8], showing that too large temporal window size brings noise to informative local representations, which also demonstrates the importance of involving locality inductive bias in early stages. 

\begin{table}[t]
    \caption{The effect of CPE when varying the number of temporal tokens on the performance (in mAP at different tIoU thresholds) and computational cost (in GFLOPs) on (a) THUMOS14 and (b) ActivityNet 1.3.}
    \label{table:act_tok_cpe}
    \begin{subtable}[h]{0.485\textwidth}
        \centering 
        \begin{tabular}{c|c|cccc}
 
            Length & GFLOPs & 0.3  & 0.4  & 0.5  & Avg. \\ \shline
            64           & 83.2  & 68.2 & 62.2 & 53.0 & 50.7 \\
            64+CPE       & 84.7  & 68.2 & 62.5 & 54.3 & 51.1 \\ \hline
            128          & 110.8 & 69.5 & 63.6 & 55.6 & 52.7 \\
            128+CPE      & 111.2 & 70.6 & 65.7 & 56.4 & 53.6 \\
             
        \end{tabular}
        \caption{THUMOS14}
        \label{table:act_tok_cpe_thumos}
    \end{subtable}
    \hfill
    \begin{subtable}[h]{0.485\textwidth}
        \centering 
        \begin{tabular}{c|c|cccc}

        Length & GFLOPs & 0.5  & 0.75 & 0.95 & Avg. \\ \shline
        192          & 181.5 & 49.8 & 32.3 & 5.3  & 32.0 \\
        192+CPE      & 182.3 & 50.9 & 33.4 & 6.9  & 33.0 \\ \hline
        96           & 133.6 & 50.9 & 33.2 & 6.1  & 32.8 \\ 
        96+CPE       & 134.1 & 51.4 & 33.7 & 6.8  & 33.4 \\
        \end{tabular}
        \caption{ActivityNet 1.3}
        \label{table:act_tok_cpe_anet}
    \end{subtable}
\end{table}
 
\noindent\textbf{Effect of CPE.} 
We verify the effectiveness of CPE in our STPT for the action detection task under the settings with different numbers of input tokens along temporal dimension. We carefully change the temporal stride in the first PatchEmbed layer, using dilated kernels to ensure more video frames are encoded into the token sequence while keeping the length of video clips identical to the setting in previous works~\cite{lin2021learning_AFSD}. Despite varying the number of input tokens, as shown in Table~\ref{table:act_tok_cpe}, CPE provides consistent performance gain on both datasets under all settings, with trivial computational cost introduced, \eg, 0.9\% average mAP improvement on THUMOS14 and 1.0\% average mAP improvement on ActivityNet 1.3 under the setting of 128 and 192, respectively. 

\section{Conclusion and Future Work}
In this paper, we have proposed a novel STPT, which tackles the challenges of both computational redundancy and long-range dependency in spatio-temporal representation learning for the task of action detection. 
Specifically, STPT applies LSTA in shallow layers to encode local patterns with reduced spatio-temporal redundancy and employs GSTA in the later stages to handle global dependencies. Extensive experiments on THUMOS14 and ActivityNet 1.3 have demonstrated that our STPT achieves a promising balance between accuracy and efficiency for the task of action detection. Future work may include extending our STPT to other dense prediction tasks in the video recognition field, \eg, video segmentation and video captioning. 

\noindent \textbf{Acknowledgment.} This work was partially supported by the NSFC under Grant (No.61972315), Shaanxi Province International Science and Technology Cooperation Program Project-Key Projects No.2022KWZ-14.

\appendix
\renewcommand\thesection{\Alph{section}}
\section*{Appendix}

We organize our supplementary material as follows. 
\begin{itemize}
    \item In Sec.~\ref{sec:lsta_illus}, we present an illustrated example of LSTA.
    \item In Sec.~\ref{sec:vis_thumos}, we compare the per-category performance of different models on THUMOS14 and provide attention visualization from the last layer on the two selected action instances.
    \item In Sec.~\ref{sec:anet_ablation}, we present additional experiment results on ActivityNet 1.3.
    \item In Sec.~\ref{sec:othervit}, we provide supplementary comparison results with other representative video Transformers.
\end{itemize}

\section{Additional Illustration of LSTA}
\label{sec:lsta_illus}
\begin{figure}
    \vspace{-2em}
    \centering
    \includegraphics[width=\textwidth]{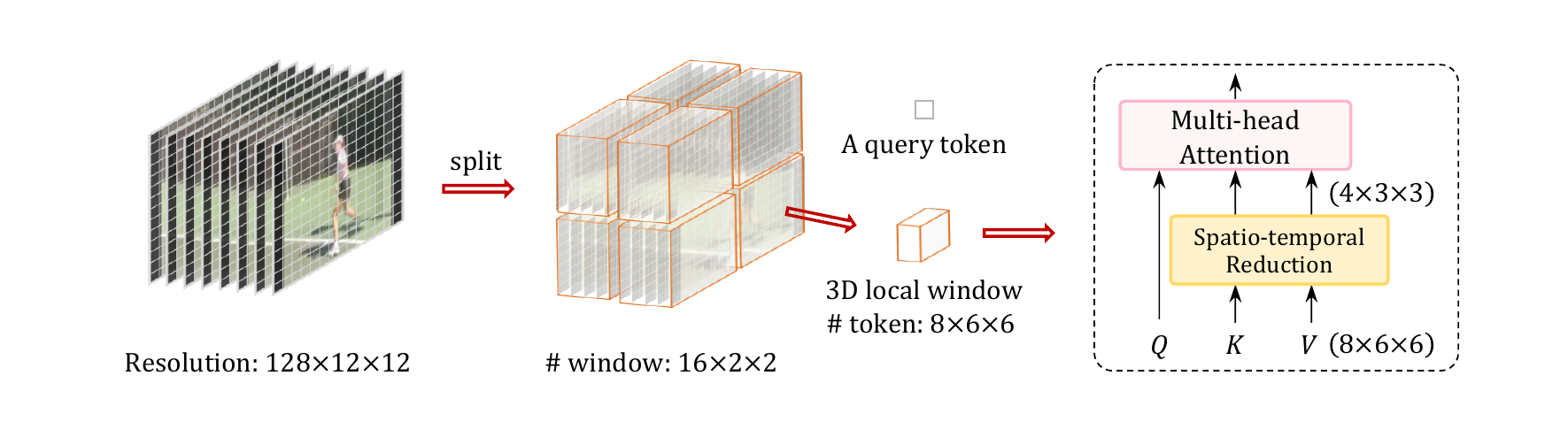}
    \caption{An illustrated example of LSTA.}
    \label{fig:local_attn}
    \vspace{-2em}
\end{figure}

We further illustrate LSTA in Fig.~\ref{fig:local_attn}. From left to right, for each input tensor of size $T\times H\times W$, we first evenly divide it into $w_1 \times w_2 \times w_3$ sub-windows, where $T$, $H$ and $W$ refer to the temporal size and height, width, respectively. Next, each query token only attends to tokens within the same 3D local window. Last, we further reduce both the spatial and temporal resolution of the keys and values within each 3D local window for better efficiency. 

\section{Attention Visualization on THUMOS14}
\label{sec:vis_thumos}
\begin{figure}[t]
    \centering
    \includegraphics[width=\textwidth]{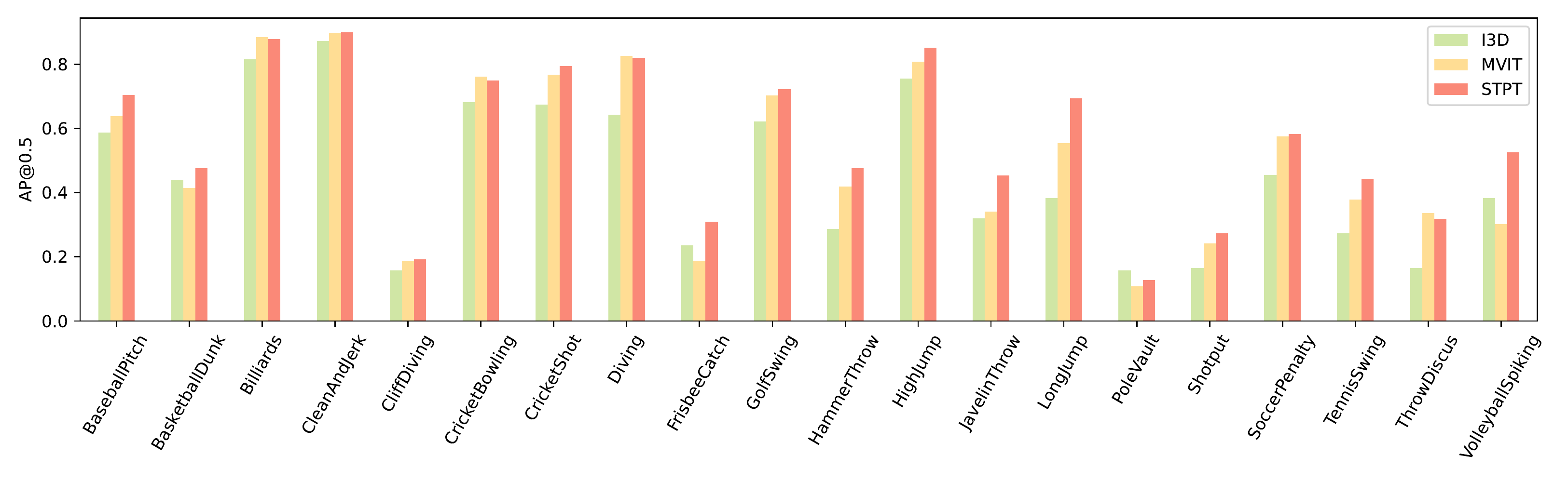}
    \caption{Per-category AP@0.5 on THUMOS14.}
    \label{fig:per_map}
\end{figure}

In Fig.~\ref{fig:per_map}, we show the per-category AP@0.5 of I3D, MViT and the proposed STPT on THUMOS14 with RGB input only. 
It demonstrates that our STPT surpasses the other two models in most categories. It is also notable that our STPT outperforms other models by a large margin on some action categories, \eg, TennisSwing, LongJump. 

\begin{figure}[!t]
\centering
\includegraphics[width=0.8\textwidth]{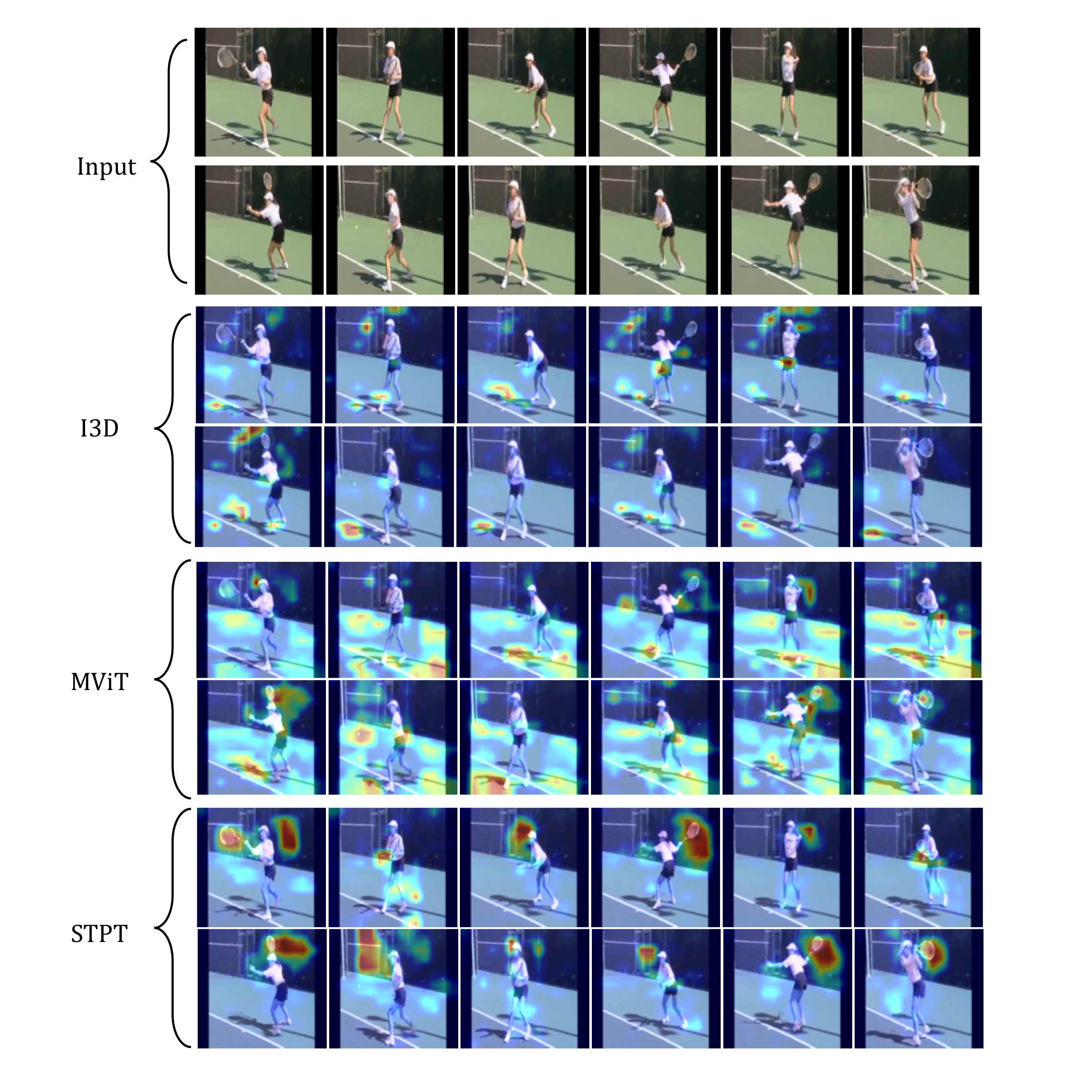}
\caption{Attention visualization of different models for an action instance of  ``TennisSwing''.}
\label{fig:v26_input+cam}
\end{figure}

\begin{figure}[!t]
\centering
\includegraphics[width=0.8\textwidth]{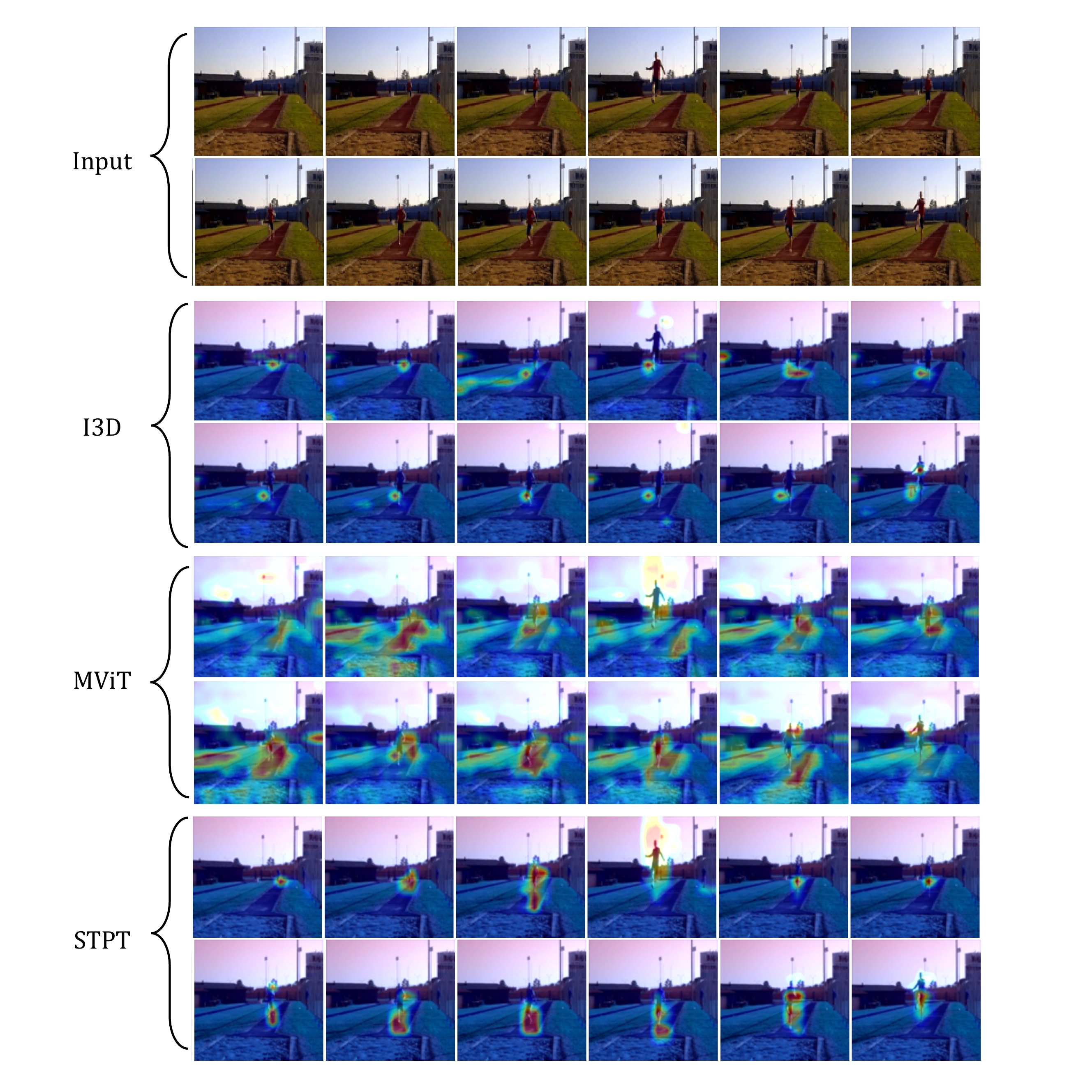}
\caption{Attention visualization of different models for an action instance of ``LongJump''. }
\label{fig:v1079_input+cam}
\end{figure}

We further provide attention visualization of different models on the two example action instances selected from THUMOS14 in Fig.~\ref{fig:v26_input+cam} and Fig.~\ref{fig:v1079_input+cam}. For each video instance, we show the sampled input RGB frames and use Grad-CAM
to generate the corresponding attention in the last layer. 
Due to the local receptive field of 3D convolutions, I3D lacks the capacity of learning long-term dependencies in the videos, leading to inaccurate and irrelevant attention. 
Alternatively, MViT applies global attention in all the stages, which brings noises to the informative spatio-temporal representations and struggles to focus on key objects or actions, \eg, the tennis racket or the jumping action. 
Different from both cases, by flexibly involving locality constraint in early stages and applying global attention in later stages, our STPT efficiently encodes local patterns and captures global dependencies in a concise manner, enabling learning strong spatio-temporal representations from videos. 

\setlength{\tabcolsep}{4pt}
\begin{table}[]
\caption{Effect of our architecture design principle. We evaluate the performance (in mAP) and computational cost (in GFLOPs) of several combinations of blocks on ActivityNet 1.3. L/G refers to LSTA/GSTA used in each stage. Models are equipped with CPE. The number of temporal tokens is set to 96. }
\label{table:anet_bt}
\centering 
\begin{tabular}{c|c|cccc}
        Type & GFLOPs & 0.5  & 0.75 & 0.95 & Avg. \\ \shline
        LLLL & 121.0  & 49.9 & 32.4 & 3.3  & 31.7 \\ \hline
        LLLG & 126.2  & 50.0 & 32.5 & 3.5  & 31.8 \\ \hline
        LGGG & 152.2  & 50.3 & 32.5 & 6.8  & 32.7 \\ \hline
        GGGG & 172.4  & 50.1 & 32.7 & 5.9  & 32.2 \\ \hline
        LLGG & 134.1  & 51.4 & 33.7 & 6.8  & 33.4
\end{tabular}
\end{table}

\setlength{\tabcolsep}{4pt}
\begin{table}[]
\caption{Effect of window size in terms of temporal dimension. We compare the performance (in mAP) and computational cost (in GFLOPs) for different scales of a local window in each LSTA block. }
\label{table:anet_ws}
\centering 
\begin{tabular}{c|c|cccc}
Window Size    & GFLOPs & 0.5  & 0.75 & 0.95 & Avg. \\ \shline
{[}1,1,1{]}    & 133.9  & 50.3 & 33.0 & 4.0  & 32.1 \\ \hline
{[}4,4,4{]}    & 134.0  & 50.4 & 33.2 & 4.1  & 32.3 \\ \hline
{[}8,8,8{]}    & 134.1  & 51.0 & 33.5 & 5.7  & 32.9 \\ \hline
{[}8,8,16{]}   & 134.1  & 51.4 & 33.7 & 6.8  & 33.4 \\ \hline
{[}16,16,16{]} & 134.4  & 49.7 & 32.0 & 5.6  & 32.0
\end{tabular}
\end{table}

\begin{table}[]
\caption{More comparisons (mAP(\%) at different tIoU thresholds) with other ViT models on THUMOS14 and  ActivityNet 1.3.}
\label{table:localglobal}
\begin{subtable}[h]{0.95\linewidth}
    \centering
        \begin{tabular}{l|c|cccccc}
        Backbone      & GFLOPs & 0.3  & 0.4  & 0.5  & 0.6  & 0.7  & Avg.  \\ \shline
        DualFormer    & 112.2  & 68.5 & 63.0 & 54.4 & 41.8 & 27.7 & 51.1  \\
        RegionViT     & 181.3  & 68.6 & 62.7 & 53.5 & 41.7 & 28.8 & 51.1  \\
        Twins         & 119.8  & 69.7 & 62.7 & 54.1 & 43.2 & 28.7 & 51.7  \\
        Uniformer & 134.2  & 68.7 & 63.6 & 54.6 & 42.0 & 28.7 & 51.5 \\
        Ours        & \bf{111.2}  & \bf{70.6} & \bf{65.7} & \bf{56.4} & \bf{44.6} & \bf{30.5} & \bf{53.6}
        \end{tabular}
    \label{table:thumos_localglobal}
    \caption{THUMOS14}
    
\end{subtable}

\begin{subtable}[h]{0.95\linewidth}
\centering
\begin{tabular}{l|c|cccc}
Backbone      & GFLOPs & 0.5  & 0.75 & 0.95 & Avg.  \\ \shline
DualFormer    & 171.1  & 50.7 & 33.1 & 5.4  & 32.7  \\
RegionViT     & 241.1  & 51.3 & 33.4 & 6.2  & 32.8  \\
Twins         & 140.2  & 51.1 & 33.2 & 5.2  & 32.7  \\
Uniformer     & 185.5  & 50.7 & 32.8 & 5.2  & 32.6  \\
Ours          & \bf{134.1}  & \bf{51.4} & \bf{33.7} & \bf{6.8}  & \bf{33.4} 
\end{tabular}
\label{table:anet_localglobal}
    \caption{ActivityNet 1.3}
\end{subtable}
\end{table}

\section{More Results on ActivityNet 1.3}
\label{sec:anet_ablation}
We provide more results on ActivityNet 1.3 in Table~\ref{table:anet_bt} and Table~\ref{table:anet_ws}. In Table~\ref{table:anet_bt}, we report the performance and the computational cost of all the possible combinations of LSTA (L) and GSTA (G) on ActivityNet 1.3. 
Without using GSTA, the model LLLL is computationally efficient but lacks the capacity of learning global dependency, resulting in the lowest mAP scores compared with other structures. Alternatively, applying GSTA in all the stages leads to heavy computational cost (134.1G \vs 172.4G). However, the mAP scores drop at all thresholds as the model cannot extract detailed spatio-temporal patterns in the early stages. Thus, for ActivityNet, we choose LSTA and GSTA in the first two stages and the last two stages respectively, in order to achieve a favourable balance between efficiency and effectiveness. 
As shown in Table~\ref{table:anet_ws}, with comparable FLOPs, the model with the window size of [8,8,16] for LSTA outperforms the other settings at all thresholds. 

\section{Comparison with other video Transformers}
\label{sec:othervit}
Compared with DualFormer, RegionViT and Twins, which alternatively process local and global information within each block, we leverage local LSTA in the early stages to remove local redundancy and utilize global GSTA in the deeper layers to model the long-term dependencies. Different from the factorized space-time attention used in RegionViT, we encode the target motions by jointly aggregating spatio-temporal relations. As shown in Table~\ref{table:localglobal}, our STPT consistently achieves better performance with fewer FLOPs than other methods on both datasets.

\bibliographystyle{splncs04}
\bibliography{egbib}

\end{document}